\documentclass[letterpaper, 10 pt, conference,hyphens]{ieeeconf}

\usepackage{cite}
\usepackage{amsmath,amssymb,amsfonts}
\usepackage{algorithmic}
\usepackage{hyperref}
\usepackage{authblk}
\usepackage{booktabs}
\usepackage{threeparttable}
\usepackage{wrapfig}
\usepackage{adjustbox}
\usepackage{comment}
\usepackage{graphicx}
\usepackage{textcomp}
\usepackage{xcolor}
\def\BibTeX{{\rm B\kern-.05em{\sc i\kern-.025em b}\kern-.08em
    T\kern-.1667em\lower.7ex\hbox{E}\kern-.125emX}}
\begin{document}

\title{\textbf{Emotional Expression in Low-Degrees-of-Freedom Robots:\\ Assessing Perception with Reachy Mini}}%How People Perceive Emotional Expression in Reachy Mini}

\author{
\begin{minipage}{0.88\textwidth}
\centering
Amit Rogel$^{1}$ \quad
Elmira Yadollahi$^{2}$ \quad
Guy Laban$^{3,4,5,*}$\\

\small
$^{1}$Robotic Musicianship Lab, Georgia Institute of Technology, Atlanta, GA, USA\\
$^{2}$School of Computing and Communications, Lancaster University, Lancaster, UK\\
$^{3}$Department of Industrial Engineering and Management, Ben-Gurion University of the Negev, Beer Sheva, Israel\\
$^{4}$School of Brain Sciences and Cognition, Ben-Gurion University of the Negev, Beer Sheva, Israel\\
$^{5}$The Azrieli National Center for Autism and Neurodevelopment Research, Ben-Gurion University of the Negev, Beer Sheva, Israel\\

\footnotesize
$^{*}$Corresponding author: \href{mailto:laban@bgu.ac.il}{\texttt{laban@bgu.ac.il}}
\end{minipage}
}

\maketitle

\begin{abstract}
Emotion expression is central to human–robot interaction, yet little is known about how people interpret affect on robots with sparse, non-anthropomorphic expressive capabilities. This study examined how people perceive emotional expressions displayed by Reachy Mini (Pollen Robotics and Hugging Face), a low-degree-of-freedom (low-DoF) robot with a constrained and distinctly non-human expressive repertoire. In an online within-subjects study, 100 participants viewed 10 short video clips of Reachy Mini expressing different emotions and, for each clip, identified the perceived emotion, rated its valence and arousal, and evaluated the robot on social-perception traits. Exact emotion recognition was modest overall and varied considerably across expressions, with anger, sadness, and interest recognized more reliably than emotions such as love, pleasure, shame, and disgust. However, participants were generally more successful at recovering broader affective meaning than exact emotion labels, particularly along valence and arousal dimensions. Emotional expressions also shaped social evaluation, as positive expressions were perceived as warmer and more sociable than negative ones, and animacy varied less across conditions. These findings suggest that even constrained robotic expressions can communicate affective meaning and influence social impressions, positioning Reachy Mini as a useful benchmark for studying affective communication in low-DoF robots.
\end{abstract}

\section{Introduction}
Emotional expression is central to social interaction, and it has become a major focus in human--robot interaction (HRI) because it shapes how robots are understood, responded to, and evaluated \cite{StockHomburg2022}. Research in this area has shown that robot expressions can influence trust, engagement, and social interpretation, while also demonstrating that people often attempt to read affective meaning even from artificial or highly simplified agents \cite{StockHomburg2022,Dubal2011,Embgen2012}. However, the literature has focused more strongly on robots with relatively rich expressive channels, such as humanoid or highly actuated platforms, than on robots whose expressive capacity is technically sparse or structurally constrained \cite{StockHomburg2022,GaoShenJiTian2024,Spitale2025PastWell-being}. As a result, we know little about how people perceive emotion when robotic expression is conveyed through a narrow and distinctly non-human repertoire. This gap is important both theoretically and practically. From a theoretical perspective, studying emotionally expressive low-degree-of-freedom robots can help clarify which aspects of emotional communication remain legible when expression differs substantially from typical human or humanoid patterns. Prior studies suggest that people can extract emotional meaning even from simplified, abstract, or non-humanoid robot behavior \cite{Dubal2011,Embgen2012,Ghafurian2022}, but recognition is not uniform across platforms or emotions, and morphological constraints may substantially shape which expressions are easier or harder to interpret \cite{Ghafurian2022,GaoShenJiTian2024}. This makes low-degree-of-freedom robots a useful test case for examining the perceptual boundaries of robot emotion expression.

Reachy Mini (Pollen Robotics and Hugging Face), a compact desktop robot with a seemingly humanoid head but no facial expressions, offers a particularly relevant case for such an enquiry. Featuring an expressive 6-DoF head, a rotating base, and animated antennae, and as a new open-source and accessible platform, it is likely to see increasing use in research and development settings. At the same time, its expressive capabilities are relatively constrained compared to more fully actuated social robots. This makes it valuable not only as a platform for future interaction design but also as a benchmark case for understanding how people perceive emotional displays from a robot whose expressive behavior is limited and clearly robot-specific. Establishing such evidence is important because HRI research benefits from systematic data on how different robot platforms are perceived, rather than assuming that intended expressions are interpreted as designed \cite{StockHomburg2022,Ghafurian2022}. More broadly, this study speaks to a wider need in HRI to build cumulative foundations for comparing emotional expression across robot morphologies and expressive architectures. Ideally, similar investigations should be conducted across many robot platforms so that the field can develop a clearer and more systematic understanding of how humans perceive and recognize robotic expressed emotion. Reachy Mini is therefore not only interesting in its own right, but also useful as a testbed or even an early benchmark within this broader agenda. In addition, robot emotional expressions may shape social impressions of the agent itself. In HRI, expressions are not only decoded as signals of internal state, but they can also influence how warm, friendly, trustworthy, or otherwise socially meaningful a robot appears \cite{Henschel2021WhatYou}. Examining these downstream social perceptions of Reachy Mini's emotional expression extends the study's contribution beyond emotion recognition alone to the broader interpersonal significance of robot expression.

Accordingly, the objective of the present study is to examine how people perceive, recognize, and interpret emotional expressions displayed by a low-DoF robot, Reachy Mini. More specifically, the study investigates the extent to which intended emotions are legible on a technically and expressively constrained robotic platform, whether participants recover broader affective meaning in terms of valence and arousal even when exact emotion recognition is weaker, and whether different emotional expressions shape social evaluations of the robot itself. Hence, we address the following research questions:

\textbf{RQ1.} \textit{To what extent do people recognize the intended emotional expressions displayed by Reachy Mini?}

\textbf{RQ2.} \textit{To what extent do Reachy Mini’s expressions communicate broader affective meaning, including valence and arousal, beyond exact emotion-label recognition?}

\textbf{RQ3.} \textit{How do Reachy Mini’s emotional expressions shape social evaluations of the robot, and how do these evaluations vary across expressed emotions?}

\begin{comment}
    
Accordingly, the objective of the present study is to examine how people perceive, recognize, and interpret emotional expressions displayed by a low-DoF robot, Reachy Mini. The study investigates the extent to which intended emotions are legible on a technically and expressively constrained robotic platform, identifies the perceptual strengths and limitations of its expressive repertoire, and assesses whether these expressions communicate broader affective meaning in terms of perceived valence and arousal. Furthermore, the study examines whether different emotional expressions shape social evaluations of the robot itself, and how such evaluations vary across expressed emotions, thereby extending the analysis beyond recognition accuracy to the wider interpersonal implications of robot emotion expression. Hence, we address the following research questions:

\textbf{RQ1.} To what extent do people recognize the intended emotional expressions displayed by Reachy Mini?

\textbf{RQ2.} Which emotions are recognized more or less reliably, and what confusion patterns characterize participants' perception of Reachy Mini's expressions?

\textbf{RQ3.} How do Reachy Mini’s emotional expressions shape social evaluations of the robot, and to what extent do different expressions differ in their social evaluations?
\end{comment}

\section{Related Works}

Research on emotion expression in human--robot interaction has shown that affect can be communicated through a range of channels, including faces, body movement, and broader motion patterns, but recognition remains strongly dependent on embodiment and study design. Stock-Homburg \cite{StockHomburg2022} reviews two decades of work and shows that the literature is broad but methodologically fragmented, making it difficult to compare findings across platforms. In parallel, Hoffman and Ju \cite{HoffmanJu2014} argue that movement should be treated as a primary expressive resource in robot design rather than as merely functional motion, and Bretan, Hoffman, and Weinberg \cite{BretanHoffmanWeinberg2015} demonstrate that dynamic physical behavior can be systematically designed to convey affective qualities through timing, energy, and kinematic variation.

These questions become especially important for low-DoF and non-anthropomorphic robots, whose restricted movement space does not allow straightforward translation of human facial or bodily expressions. Ju and Takayama \cite{JuTakayama2009} showed that, even outside traditional social robotics, automatic door motion can be interpreted as a socially meaningful gesture. Closely related work by Anderson-Bashan et al. \cite{AndersonBashan2018GreetingMachine} demonstrated that an abstract robotic object could communicate positive and negative social cues in opening encounters. Building on that line, Erel, Hoffman, and Zuckerman \cite{ErelHoffmanZuckerman2018Gestures} reported that gestures of non-anthropomorphic robots can be interpreted as conveying both social intent and emotion, despite severe mechanical simplicity. Importantly, in their study with a 2-DoF robot, vertical movement and approach-versus-avoid direction emerged as especially salient dimensions, with upward and toward-participant motion associated with more positive valence and downward or away motion associated with more negative valence. Erel et al. \cite{ErelShemTovKesslerZuckerman2019} showed that abstract robot movement is interpreted as a social cue, even automatically: in an Implicit Association Test, participants were slower when movement direction conflicted with socially congruent meanings, suggesting that social interpretation was activated even when it hindered task performance. Rifinski et al. \cite{RifinskiErelFeinerHoffmanZuckerman2020} further showed that minimal responsive behaviors by a non-humanoid robotic object, specifically gaze and leaning, altered interpersonal evaluation in human--human conversation. Rogel et al. used animation techniques to show that slight modifications to a robotic arm's trajectory would improve its perceived animacy social scores \cite{rogel2022robogroove}. Together, these studies suggest that sparse and robot-specific movement can shape both affective interpretation and social evaluation, even in the absence of humanoid embodiment.

%A closely related line of work has examined how simple, non-humanoid robots can communicate socially meaningful states through sparse motion. %Building on earlier studies on abstract robotic objects and minimal responsive behavior \cite{AndersonBashan2018GreetingMachine,ErelHoffmanZuckerman2018GesturesGestures,RifinskiErelFeinerHoffmanZuckerman2020,ErelEtAl2022Support}, 
More recent work has extended this space toward both new expressive functions and novel generation methods. For instance, Press and Erel \cite{PressErel2023Humorous} showed that humorous non-verbal behavior by a non-humanoid robot can reduce social awkwardness. Extending this line of work, Vidra et al.~\cite{VidraEtAl2025GREG} proposed a transfer-learning approach for generating non-verbal robotic social gestures from a small seed set of animations. Suguitan et al.\ \cite{SuguitanDePalmaHoffmanHodgins2024Face2Gesture} demonstrated on the Blossom platform \cite{SuguitanHoffman2019Blossom} that affective robot movement can be generated from facial-expression priors. Together, these findings support the idea that emotional meaning can be conveyed through robot-specific movement patterns rather than only through human-like facial reproduction.
%Recent studies have extended this literature, but many 
Other studies focus either on richer embodiments or on downstream interaction effects, rather than on highly constrained, open, and widely accessible platforms. Gao et al. \cite{GaoShenJiTian2024} examined the perception of humanoid robot body movements and showed that participants evaluate robot expressions in terms of emotion category, intensity, and arousal. De Rooij et al. \cite{deRooij2024} found that robot mood expressions can influence perceived valence and arousal and, through these perceptions, affect collaboration, satisfaction, and performance in co-creative group settings. Fujii et al. \cite{Fujii2025} further showed that even advanced android facial expressions vary in perceived emotional readability depending on how participants are instructed and what interpretive frame they bring to the task. More broadly, emotional perception in artificial agents is inherently multimodal, integrating visual and auditory cues that may be limited on constrained platforms \cite{torre2022smiling}. Thus, these studies suggest that the field has made recent progress in understanding robot emotion expression, but also highlight a gap: the lack of systematic benchmarks on more constrained platforms, where expressions are likely to be readable in some respects yet limited in others.

This is where Reachy Mini becomes analytically useful: it allows the study of what emotional information remains legible when expression is technically sparse, robot-specific, and not supported by a full humanoid expressive repertoire. Prior work involved creating a set of frameworks for generating emotional gestures on various non-anthropomorphic robots. These frameworks mapped specific human emotional gestures to various degrees of freedom in robotic arms and mobile manipulators, such as the stretch. Accompanying these gestures were generated emotional musical phrases for various emotions along the Geneva Emotion Wheel \cite{savery2021emotional}. These emotional phrases were evaluated to understand the effective communication of affect through coordinated bodily movement, gesture design, and prosodic modulation across different robotic platforms \cite{savery2023robotic}. This methodology was designed to apply to a broad scope of robots.

Reachy Mini itself has only recently begun to appear in the scholarly literature, and so far mainly as a platform reference rather than as a primary object of empirical study. Vonschallen et al.~\cite{VonschallenEtAl2026} used Reachy Mini as an illustrative case in a study on design requirements for generative social robots in higher education, while Antony et al.~\cite{AntonyEtAl2026IntroducingM} cite Reachy Mini as an example of a compact, accessible platform oriented toward AI experimentation and head-based expression. Similarly, \cite{AldarondoEtAl2026FaunaSprout} situates Reachy Mini within the emerging design space of lightweight, expressive small-scale robots, in a work addressing Fauna, an expressive humanoid. To the best of our knowledge, however, no prior published or arxived work has systematically examined how people perceive and recognize Reachy Mini's emotional expressions, or how they attribute social meaning to them. This makes Reachy Mini a timely and relevant platform for establishing foundational evidence on the legibility and social interpretation of emotion expression in a low-DoF robot.

Beyond recognition accuracy, emotional expressions may also shape how a robot is evaluated as a social other. This broader interpersonal relevance can be understood through the Emotions as Social Information (EASI) model \cite{VanKleef2009EASI}. The model argues that emotional expressions influence observers not only by signaling an internal state, but also by eliciting affective reactions and prompting inferential processes about the expresser's intentions, orientation, and likely behavior \cite{VanKleef2009EASI,VanKleefCote2022}. %Although EASI originates in social psychology rather than HRI, its core logic is highly relevant here and was previously addressed in the HRI context~\cite{Laban2024SharingFeel}. 
From this perspective, robot expressions matter not only in terms of whether people can identify the intended emotion, but also in how these expressions shape impressions of the robot itself. Emerging HRI findings are consistent with this view. Song et al.\ \cite{SongTaoLuximon2023} found that positive rather than negative robot emotional expressions increased perceived anthropomorphic trustworthiness, and de Rooij et al.\ \cite{deRooij2024} showed that robot mood expressions influenced participants' responses through perceived robot valence and arousal. Laban et al.\ \cite{LabanGeorgeMorrisonCross2021TellMeMore} showed that embodiment influences the quantity and quality of self-disclosure to artificial agents, as well as their social perception of the robot. More recent work has demonstrated that repeated interactions with a social robot can enhance users' emotional understanding and sense of control, alongside changes in emotional expressiveness over time \cite{LabanWangGunes2026EmotionRegulation}. These findings suggest that robot affect is consequential not only as a perceptual signal, but also as a factor shaping the robot's social meaning and the impressions people form during an interaction. In the context of the present study, this perspective motivates examining whether Reachy Mini's expressed emotions differ not only in recognizability, but also in the social evaluations they elicit.%, such as warmth, friendliness, trustworthiness, and related impressions.

\section{Methods}

\subsection{Participants and Design}

The study employed a within-subjects online experimental design in which each participant evaluated the full set of Reachy Mini emotional-expression stimuli. An a priori power analysis was conducted for a within-subject comparison using a two-tailed paired-samples test with $\alpha=.05$, power = .80, and a small-to-moderate expected effect size of $d_z=.30$. The analysis indicated a sample of approximately 90 participants was required. Accordingly, a total of 100 participants were recruited via Prolific from the United States and the United Kingdom, with potential dropouts taken into account. Eligible participants had to be at least 18 years old, able to complete the study in English, reside in the US or UK, have corrected or fixed vision and hearing, have no diagnosis or self-diagnosis of autism, and be able to complete the study from their personal computer. Participants completed the study on their own computers and were instructed to use speakers or headphones throughout the task. Compensation was set at £1.50 per 10 minutes of participation (£9 per hour), 1 minute per treatment. The final sample included 100 participants, primarily identifying as female (59\%) or male (40\%), with one participant identifying as non-binary, and was almost evenly split between the United States (51\%) and the United Kingdom (49\%). Participants had an approximate mean age of 48.53 years ($SD = 13.07$, range = 19--80).

\subsection{Stimuli and Procedure}

The stimuli consisted of short video clips of Reachy Mini displaying emotional expressions. Each clip lasted approximately 10 seconds and included audio. Reachy Mini emotional expression gestures were created using Pollen Robotics model for emotion generation on the robot\footnote{\url{https://huggingface.co/spaces/RemiFabre/feeling_machine}}, a standardized demo feature accessible through their open-source platform\footnote{\url{https://github.com/InTheSnow31/reachy-mini-I3R/tree/main/docs/report}\label{fn:github}}. %A Large Language Model (LLM)\footnote{\url{https://github.com/InTheSnow31/reachy-mini-I3R/tree/main/docs/report}\label{fn:github}} prompted with 
They mapped Laban Movement Analysis \cite{groff1995laban} %mapped 
movements across the robot's available DoFs to represent different levels of valence and arousal\cite{durand2026reachy}. The target valence and arousal values for each emotion were derived from established affective models \cite{russell1977evidence,sacharin2012geneva}. In addition to movement, the robot used non-verbal audio cues\cite{durand2026reachy} %were generated using a pretrained audio generation model from Hugging Face\footref{fn:github}
, found in the same repository\footref{fn:github}. Using the audio and gestural model, 10 emotional gestures were created: 5 emotions with positive valence, and 5 with negative valence.

\begin{figure}[!t]
    \centering
    \includegraphics[width=0.85\columnwidth]{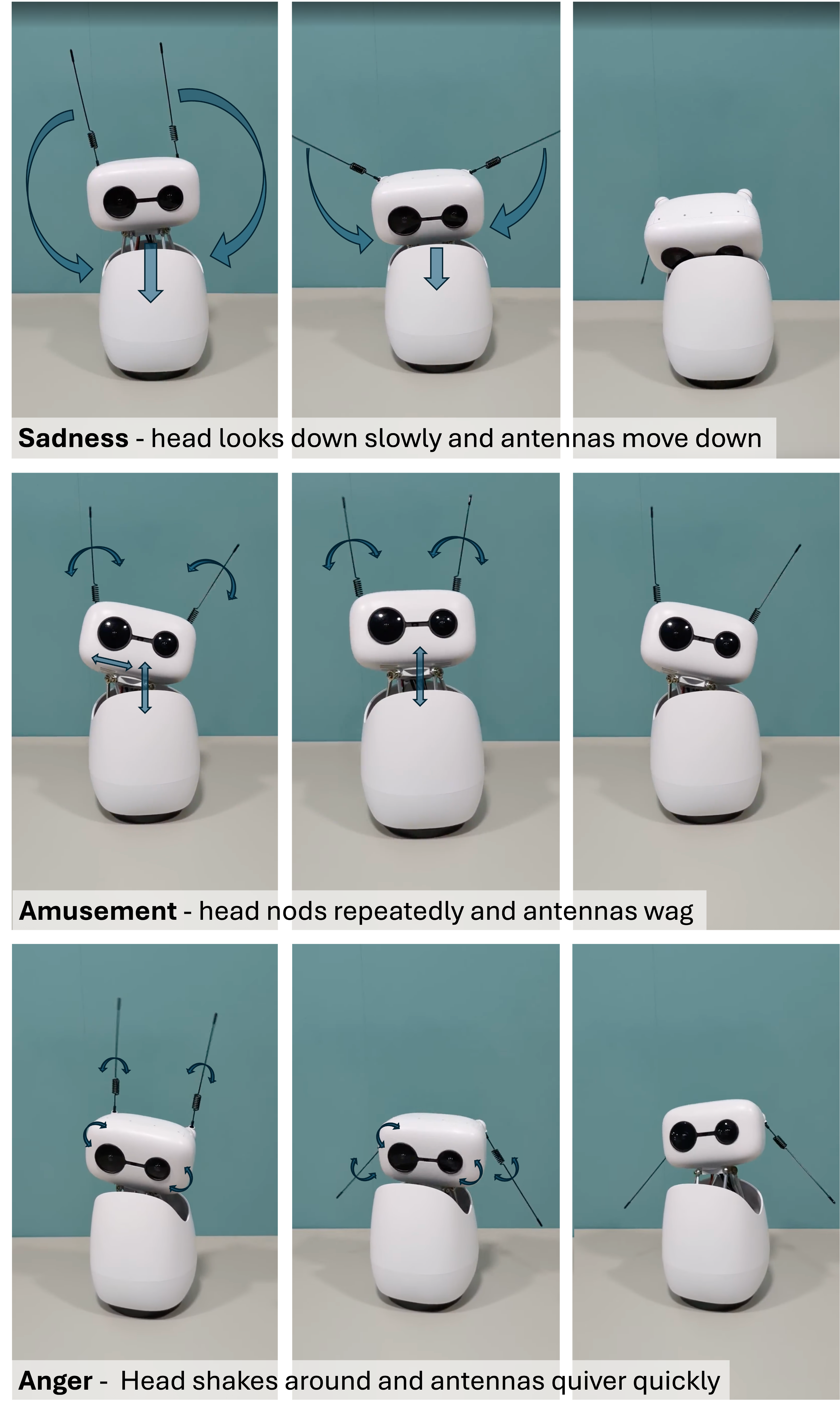}
    \caption{\footnotesize The gestures of Reachy Mini for amusement, sadness, and anger.}
    \label{fig:emotion}
    \vspace{-2mm}
\end{figure}

Participants completed the study in a single session. On each trial, they viewed one clip, selected the emotion they believed the robot was attempting to portray, rated the perceived valence and arousal of the expression, and evaluated the robot using a set of social-perception descriptors. The task took approximately 10 minutes to complete. The study used a repeated-measures design, in which each participant viewed all target expressions for Shame, Joy, Fear, Amusement, Disgust, Anger, Sadness, Love, Pleasure, and Interest.

\subsection{Measures}

\subsubsection{Emotion recognition}
Emotion recognition was measured using a wheel-based response format adapted from the Geneva Emotion Wheel (GEW) \cite{sacharin2012geneva,scherer2013grid}. The GEW is a theoretically derived and empirically tested self-report instrument designed to assess emotional reactions to objects, events, and situations. In its standard form, it arranges 20 distinct emotion families in a circular layout and represents intensity through five graded response options, along with central “None” and “Other” options. In the present study, participants selected a single circle corresponding to the emotion family and intensity that best described what they believed the robot was trying to portray. Thus, the task used the GEW format as a structured recognition interface for perceived robot emotion, rather than as a full multi-emotion self-report measure.

\subsubsection{Dimensional affect ratings}
After each emotion-selection response, participants rated the perceived valence--arousal (VA) values of the expression on continuous scales ranging from 0 to 100. Valence was anchored from negative to positive, and arousal from passive to intense. These ratings were included to capture broader affective meaning beyond discrete category selection, allowing the analysis to test whether expressions communicated the intended affective profile even when participants did not converge on the exact same emotion label.

\subsubsection{Social perception}
Social evaluation was assessed using a brief set of descriptors centered on the robot’s perceived social warmth. This part of the survey was informed by the Robotic Social Attributes Scale (RoSAS)~\cite{carpinella2017rosas}, an 18-item standardized measure of robot social perception comprising the dimensions of \textit{warmth}, \textit{competence}, and \textit{discomfort}. In the present study, warmth-related social evaluation was captured using the descriptors warm, likeable, trustworthy, and friendly. The Human Robot Interaction Evaluation Scale (HRIES) was also used to measure sociability and animacy \cite{spatola2021perception}. The items for these measures included \textit{alive}, \textit{natural}, \textit{real}, and \textit{humanlike}. Participants rated each item on a 7-point scale ranging from 1 (Not at all) to 7 (Totally).

\subsubsection{Demographic and open-ended responses}
At the end of the study, participants provided demographic information, including country of residence, year of birth, and identified gender. They were also given the opportunity to provide open-ended comments on the experiment, the robot, and its expressions.% the extent to which the robot’s expressions were easy or difficult to recognize.

\subsection{Analysis}
%Emotion recognition was analyzed by evaluating participants' ability to identify the proper quadrant of the emotion. This method was chosen over explicit emotional accuracy, as linguistic differences may prevent participants from selecting the exact intended emotion using the GEW terms despite correctly interpreting the underlying affective meaning \cite{sacharin2012geneva,scherer2013grid}. Quadrant analysis can determine whether users correctly interpreted the sentiment conveyed by Reachy Mini's expression. The quadrants were divided into combinations of Positive/Negative valence and High/Low arousal. This allowed us to assess whether participants captured the intended affective profile of each expression, even when discrete emotion recognition varied. 

Emotion recognition was analyzed at multiple levels of specificity. We first assessed \textit{exact recognition}, defined as whether participants selected the intended emotion label. However, because linguistic or semantic differences may prevent participants from choosing the exact Geneva Emotion Wheel (\textit{GEW}) term despite correctly interpreting the broader affective meaning of the expression \cite{sacharin2012geneva,scherer2013grid}, we additionally evaluated recognition at the level of affective structure. Specifically, intended and selected emotions were mapped onto one of four quadrants defined by valence (positive vs.\ negative) and arousal (high vs.\ low): positive-high, positive-low, negative-high, and negative-low. Quadrant accuracy was defined as whether the selected emotion fell within the same VA quadrant as the intended emotion, even if the exact label differed, allowing us to test whether participants captured the intended affective profile conveyed by Reachy Mini's expression. From the same mapping, we also derived valence accuracy and arousal accuracy separately. Valence and arousal slider ratings were analyzed in parallel by dichotomizing each dimension at the midpoint of the scale and combining them into a VA-based quadrant classification for each trial. Across these measures, inferential analyses were conducted at the trial level using models that accounted for the repeated-measures structure of the data by clustering standard errors at the participant level. We used participant-clustered generalized linear models to test whether exact, quadrant-, valence-, and arousal-level accuracy differed across intended expressed emotions and to compare categorical emotion selections with VA-based classifications. For social evaluation, we constructed composite indices of sociability (warm, likeable, trustworthy, and friendly; $M$ = 4.24, SD = 1.32, $\alpha$ = .91) and animacy (alive, natural, real, and humanlike; $M$ = 3.26, $SD$ = 1.43, $\alpha$ = .90), while also retaining warmth as a single-item outcome ($M$ = 4.03, SD = 1.57). We then used participant-clustered linear models to test whether intended expressed emotion and broader valence category predicted social evaluation, supplemented by Pearson correlations and generalized estimating equation (GEE) models examining associations between continuous VA ratings and social-trait judgments. Unless otherwise noted, tests were two-tailed with $\alpha = .05$.%, and effect estimates are reported with 95\% CIs where applicable.

% \section{Results}

\section{Results}

\subsection{Emotion recognition}

\begin{figure*}[h!]
    \centering
    \includegraphics[width=\textwidth]{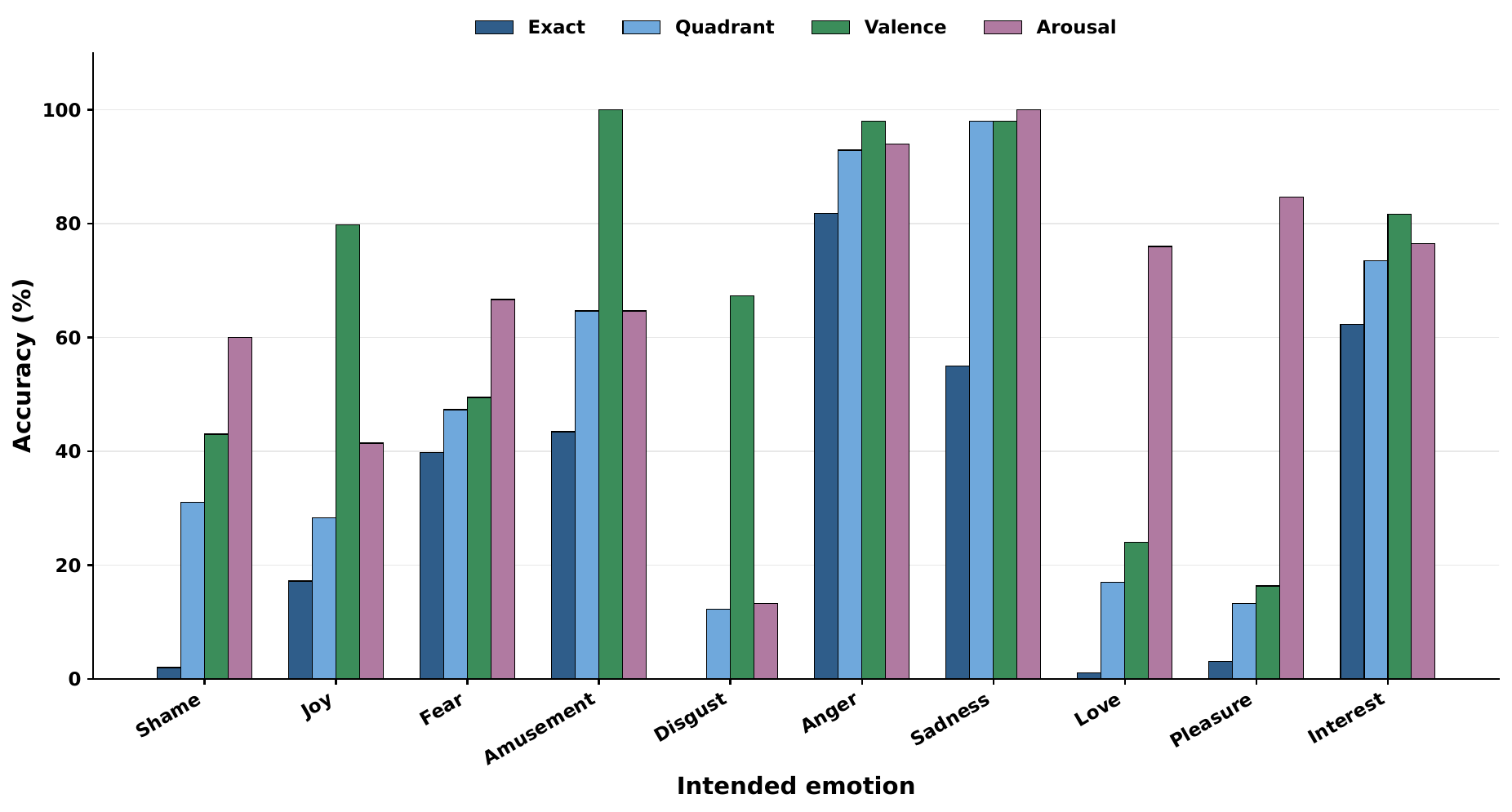}
    \caption{Recognition accuracy by intended emotion, showing label recognition and broader affective recovery at the quadrant, valence, and arousal levels.}
    \label{fig:fig1}
\end{figure*}

A participant-clustered logistic model showed that intended emotion significantly predicted exact recognition, $\chi^2(9) = 439.79$, $p < .001$. Overall exact-label accuracy was $30.5\%$. Recognition was highest for \textit{Anger} ($81.8\%$), \textit{Interest} ($62.2\%$), \textit{Sadness} ($55.0\%$), \textit{Amusement} ($43.4\%$), and \textit{Fear} ($39.8\%$), and lowest for \textit{Disgust} ($0.0\%$), \textit{Love} ($1.0\%$), \textit{Shame} ($2.0\%$), and \textit{Pleasure} ($3.1\%$). Thus, discrete emotion identification varied substantially across the robot's expressions. When categorical responses were collapsed into broader affective categories based on the selected label, intended emotion remained a significant predictor of quadrant match, $\chi^2(9) = 438.95$, $p < .001$, valence match, $\chi^2(9) = 444.34$, $p < .001$, and arousal match, $\chi^2(9) = 298.05$, $p < .001$. Overall choice-based accuracy was $47.9\%$ for quadrant, $65.9\%$ for valence, and $67.8\%$ for arousal.
Together, these results indicate that while discrete emotion recognition was limited, participants were more successful in recovering the broader affective structure of expressions, particularly along the valence and arousal dimensions. 

\begin{table*}[t]
\caption{Recognition accuracy by emotion.} %Percentages are reported separately for categorical emotion-choice responses and  VA slider-derived classifications.}
\label{tab:recognition_accuracy}
\centering
\footnotesize
\resizebox{.9\linewidth}{!}{%
\begin{tabular}{llccccccccc}
\toprule
& & \multicolumn{5}{c}{Choice-based accuracy (\%)} & \multicolumn{4}{c}{VA-based accuracy (\%)} \\
\cmidrule(lr){3-7} \cmidrule(lr){8-11}
\textbf{Emotion} & \textbf{Quadrant} & \textbf{$n$} & \textbf{Exact} & \textbf{Quadrant} & \textbf{Valence} & \textbf{Arousal} & \textbf{$n$} & \textbf{Quadrant} & \textbf{Valence} & \textbf{Arousal} \\
\midrule
Shame & Neg-Low & 100 & 2.0 & 31.0 & 43.0 & 60.0 & 100 & 16.0 & 35.0 & 28.0 \\
Joy & Pos-High & 99 & 17.2 & 28.3 & 79.8 & 41.4 & 100 & 74.0 & 84.0 & 84.0 \\
Fear & Neg-High & 93 & 39.8 & 47.3 & 49.5 & 66.7 & 100 & 28.0 & 36.0 & 84.0 \\
Amusement & Pos-High & 99 & 43.4 & 64.6 & 100.0 & 64.6 & 100 & 90.0 & 97.0 & 91.0 \\
Disgust & Neg-High & 98 & 0.0 & 12.2 & 67.3 & 13.3 & 100 & 31.0 & 60.0 & 64.0 \\
Anger & Neg-High & 99 & 81.8 & 92.9 & 98.0 & 93.9 & 100 & 71.0 & 80.0 & 90.0 \\
Sadness & Neg-Low & 100 & 55.0 & 98.0 & 98.0 & 100.0 & 100 & 17.0 & 76.0 & 19.0 \\
Love & Pos-Low & 100 & 1.0 & 17.0 & 24.0 & 76.0 & 100 & 7.0 & 34.0 & 45.0 \\
Pleasure & Pos-Low & 98 & 3.1 & 13.3 & 16.3 & 84.7 & 100 & 6.0 & 35.0 & 38.0 \\
Interest & Pos-Low & 98 & 62.2 & 73.5 & 81.6 & 76.5 & 100 & 6.0 & 81.0 & 14.0 \\
\midrule
\textbf{Overall} & --- & 984 & 30.5 & 47.9 & 65.9 & 67.8 & 1000 & 34.6 & 61.8 & 55.7 \\
\bottomrule
\end{tabular}%
}
\begin{flushleft}
\footnotesize Note. Choice-based analyses exclude trials in which participants selected \textit{None} or \textit{Other}, yielding $n = 984$ valid categorical responses overall. VA-based analyses include all 1000 trials. %Quadrant assignments followed the GEW mapping, with \textit{Pleasure} coded as positive-low arousal.
\end{flushleft}
\end{table*}

For the choice-based quadrant classification, performance varied strongly across intended expressions. The highest quadrant accuracy was observed for \textit{Sadness} (98.0\%) and \textit{Anger} (92.9\%), followed by \textit{Interest} (73.5\%), \textit{Amusement} (64.6\%), \textit{Fear} (47.3\%), \textit{Shame} (31.0\%), and \textit{Joy} (28.3\%). By contrast, \textit{Love} (17.0\%), \textit{Pleasure} (13.3\%), and \textit{Disgust} (12.2\%) showed poor quadrant-level recovery. These findings indicate that for some expressions, especially \textit{Sadness} and \textit{Anger}, participants were able to recover the broad affective region even when they did not select the exact intended label, reinforcing the robustness of these expressions at the dimensional level.
%For the choice-based quadrant classification, performance varied strongly across intended expressions. The highest quadrant accuracy was observed for \textit{Sadness} ($98.0\%$) and \textit{Anger} ($92.9\%$), followed by \textit{Amusement} ($79.8\%$), \textit{Interest} ($70.4\%$), \textit{Fear} ($58.2\%$), and \textit{Joy} ($50.5\%$). By contrast, \textit{Shame} ($40.0\%$), \textit{Love} ($17.0\%$), \textit{Pleasure} ($13.3\%$), and \textit{Disgust} ($12.2\%$) showed poor quadrant-level recovery. These findings indicate that for some expressions, especially \textit{Sadness} and \textit{Anger}, participants were able to recover the broad affective region even when they did not select the exact intended label, reinforcing the robustness of these expressions at the dimensional level.
% often recovered the broader affective region even when they did not select the exact intended label.

Among exact-correct responses ($n = 300$), $48.3\%$ were identified at the maximum selected intensity and $51.7\%$ at reduced intensity levels. This suggests that successful recognition did not depend exclusively on peak-intensity perception, and that participants could identify emotions even when they were perceived as less strongly expressed. 
% indicating that successful recognition did not depend exclusively on full-intensity selections.

\subsection{Categorical versus valence--arousal-based affect recovery}

When participants’ valence and arousal slider ratings were converted into affective categories using the midpoint of the scale, intended emotion significantly predicted VA-based quadrant classification, $\chi^2(9) = 426.78$, $p < .001$, VA-based valence classification, $\chi^2(9) = 255.17$, $p < .001$, and VA-based arousal classification, $\chi^2(9) = 373.93$, $p < .001$. Across all trials, VA-based accuracy was 34.6\% for quadrant, 61.8\% for valence, and 55.7\% for arousal. VA-based quadrant accuracy also varied across intended emotions. It was highest for \textit{Amusement} (90.0\%), \textit{Joy} (74.0\%), and \textit{Anger} (71.0\%), intermediate for \textit{Disgust} (31.0\%), \textit{Fear} (28.0\%), \textit{Sadness} (17.0\%), and \textit{Shame} (16.0\%), and lowest for \textit{Love} (7.0\%), \textit{Interest} (6.0\%), and \textit{Pleasure} (6.0\%). Overall, VA ratings captured the broad affective placement of some expressions, but did so less consistently than the categorical emotion choices.
%When participants' valence and arousal slider ratings were converted into affective categories using the midpoint of the scale, intended emotion significantly predicted VA-based quadrant classification, $\chi^2(9) = 426.78$, $p < .001$, VA-based valence classification, $\chi^2(9) = 255.17$, $p < .001$, and VA-based arousal classification, $\chi^2(9) = 373.93$, $p < .001$. Across all trials, VA-based accuracy was $34.6\%$ for quadrant, $61.8\%$ for valence, and $55.7\%$ for arousal. VA-based quadrant accuracy also varied across intended emotions. It was highest for \textit{Joy} ($74.0\%$), \textit{Anger} ($61.0\%$), and \textit{Amusement} ($57.0\%$), intermediate for \textit{Fear} ($41.0\%$), \textit{Love} ($38.0\%$), \textit{Interest} ($36.0\%$), \textit{Sadness} ($31.0\%$), and \textit{Shame} ($29.0\%$), and lowest for \textit{Pleasure} ($27.0\%$) and \textit{Disgust} ($23.0\%$). Overall, VA ratings captured the broad affective placement of some expressions, but did so less consistently than the categorical emotion choices. 
To directly compare the two response formats on the same valid-choice trials, we fit participant-clustered logistic models including source (categorical choice vs.\ VA-derived classification), intended emotion, and their interaction. Categorical responses yielded higher overall accuracy than VA-derived classification for quadrant ($47.9\%$ vs.\ $35.0\%$), $\chi^2(1) = 11.07$, $p < .001$, valence ($65.9\%$ vs.\ $62.2\%$), $\chi^2(1) = 4.65$, $p = .031$, and arousal ($67.8\%$ vs.\ $55.8\%$), $\chi^2(1) = 29.05$, $p < .001$. However, these differences depended strongly on the intended emotion, as shown by significant source $\times$ emotion interactions for quadrant, $\chi^2(9) = 189.42$, $p < .001$, valence, $\chi^2(9) = 1644.17$, $p < .001$, and arousal, $\chi^2(9) = 8427.30$, $p < .001$. In other words, this indicates that the relative effectiveness of categorical versus VA-based responses depends strongly on the specific emotion being expressed.

\begin{figure}[h!]
    \centering
    \includegraphics[width=\linewidth]{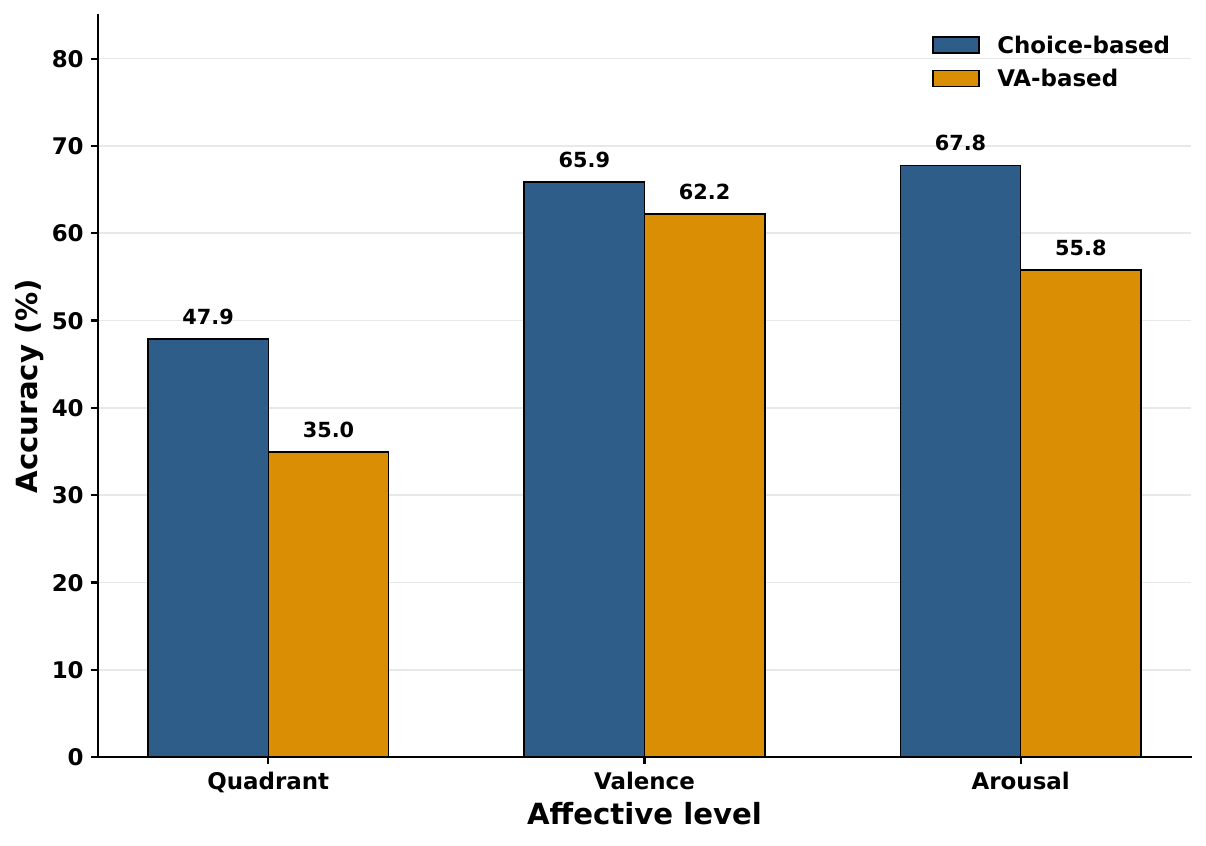}
\caption{Overall affective recovery from categorical emotion choices and VA ratings at the quadrant, valence, and arousal levels.}
    \label{fig:placeholder}
\end{figure}

\subsection{Social evaluation of the robot}

Intended emotion significantly affected sociability, $\chi^2(9) = 373.84$, $p < .001$, warmth, $\chi^2(9) = 367.00$, $p < .001$, and animacy, $\chi^2(9) = 51.63$, $p < .001$. The strongest effects emerged for sociability and warmth. \textit{Amusement} received the highest ratings on both sociability ($M = 5.07$) and warmth ($M = 5.20$), whereas \textit{Anger} received the lowest ratings on sociability ($M = 2.66$) and warmth ($M = 2.29$). Animacy varied less across expressions, ranging from $M = 3.02$ for \textit{Love} to $M = 3.44$ for both \textit{Amusement} and \textit{Sadness}$.$ 

\begin{figure*}[h!]
    \centering
    \includegraphics[width=\linewidth]{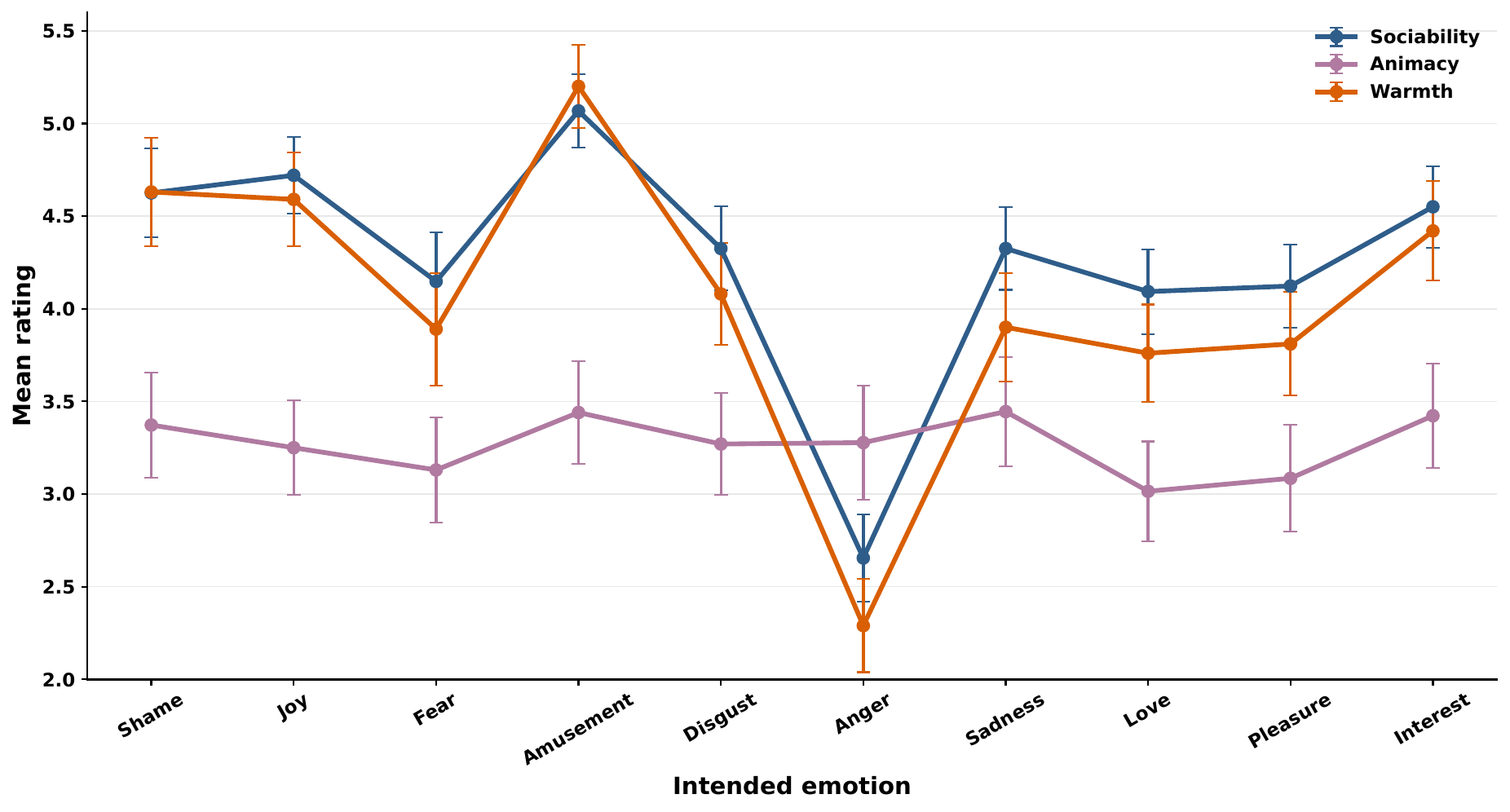}
\caption{Social evaluation across intended emotional expressions, showing mean ratings of sociability, animacy, and warmth.}
    \label{fig:fig3}
\end{figure*}

When intended emotions were grouped by valence, positive expressions were rated as more sociable than negative expressions, $b = 0.50$, 95\% CI $[0.40, 0.59]$, $p < .001$, and warmer, $b = 0.60$, 95\% CI $[0.47, 0.73]$, $p < .001$. By contrast, animacy did not differ significantly between positive and negative expressions, $b = -0.06$, 95\% CI $[-0.17, 0.06]$, $p = .339$. This pattern indicates that emotional expressions primarily shaped how socially positive the robot was perceived to be, rather than how animate it appeared.

\begin{figure}[h!]
    \centering
    \includegraphics[width=\linewidth]{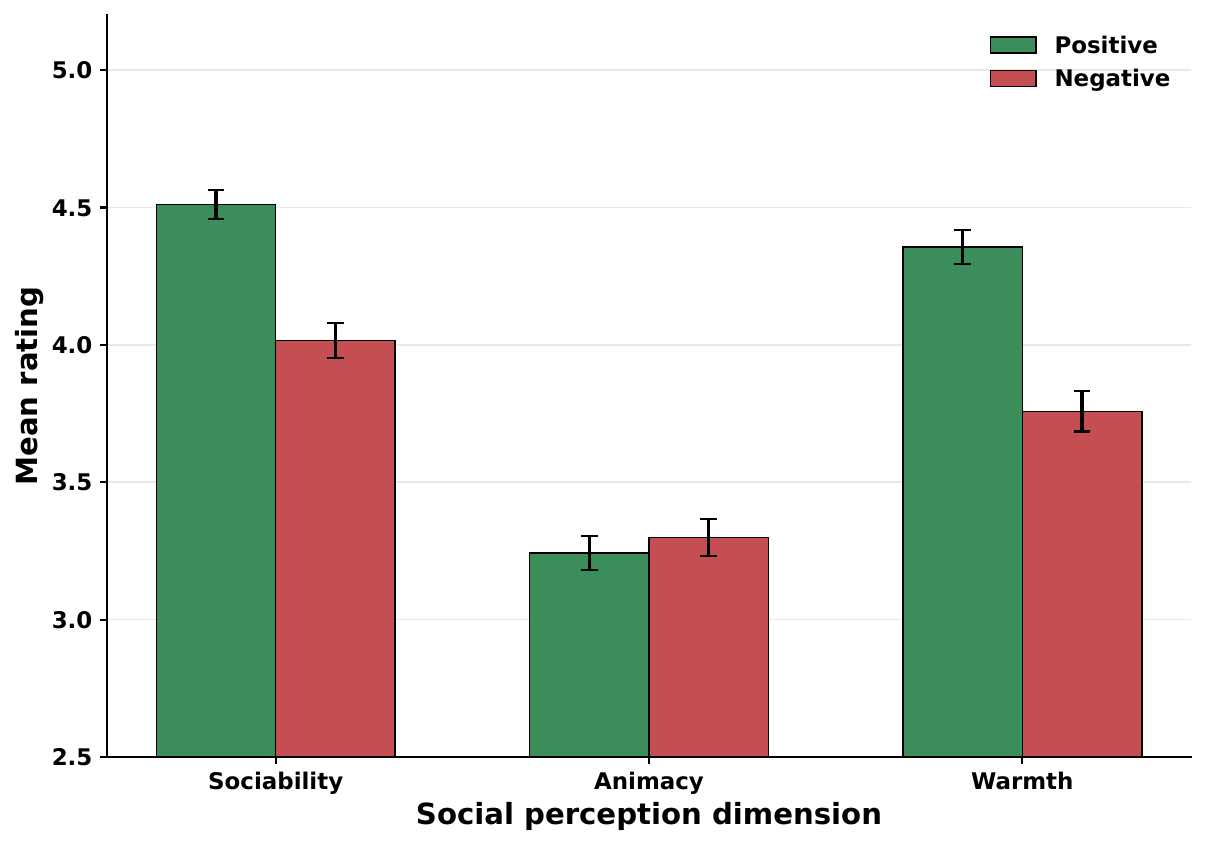}
\caption{Social evaluation of positive versus negative expressions, showing mean ratings of sociability, animacy, and warmth by valence group.}
    \label{fig:fig4}
\end{figure}

\subsection{Associations between social trait ratings and valence--arousal values}

To examine how individual social-trait ratings related to the continuous affective dimensions of valence and arousal, we first computed Pearson correlations across all 1,000 observations and then estimated GEE models with participant treated as the clustering variable. In the GEE models, valence and arousal were entered simultaneously, and coefficients were scaled such that each $\textit{b}$ reflects the expected change in the social-trait rating associated with a 10-point increase in valence or arousal.

\begin{table}[t]
\centering
\caption{Associations of social-trait ratings with valence and arousal.}
\label{tab:va_associations}
\begin{threeparttable}
\footnotesize
\begin{adjustbox}{max width=\linewidth}
\begin{tabular}{lcccccccc}
\toprule
& \multicolumn{4}{c}{\textbf{Pearson correlations}} & \multicolumn{4}{c}{\textbf{Clustered GEE coefficients}} \\
\cmidrule(lr){2-5} \cmidrule(lr){6-9}
\textbf{Trait} & \textbf{$r$ with V} & \textbf{$p$} & \textbf{$r$ with A} & \textbf{$p$} & \textbf{$b$ for V} & \textbf{$p$} & \textbf{$b$ for A} & \textbf{$p$} \\
\midrule
Warm         & .57 & $< .001$ & .15 & $< .001$ & 0.32 & $< .001$ & -0.03 & .282 \\
Likeable     & .42 & $< .001$ & .17 & $< .001$ & 0.21 & $< .001$ & -0.02 & .583 \\
Trustworthy  & .31 & $< .001$ & .22 & $< .001$ & 0.12 & $< .001$ & 0.02  & .292 \\
Friendly     & .48 & $< .001$ & .16 & $< .001$ & 0.28 & $< .001$ & -0.04 & .135 \\
Alive        & .13 & $< .001$ & .21 & $< .001$ & 0.05 & .001      & 0.14  & $< .001$ \\
Natural      & .12 & $< .001$ & .12 & $< .001$ & 0.02 & .124      & 0.10  & $< .001$ \\
Real         & .10 & .001      & .20 & $< .001$ & 0.02 & .166      & 0.10  & $< .001$ \\
Humanlike    & .08 & .011      & .19 & $< .001$ & 0.02 & .200      & 0.11  & $< .001$ \\
\bottomrule
\end{tabular}
\end{adjustbox}

\vspace{2pt}
\begin{minipage}{0.98\linewidth}
\footnotesize
\textit{Note}. \footnotesize{V = valence; A = arousal. Pearson correlations were computed across all 1,000 observations. %GEE models included valence and arousal simultaneously and treated participants as the clustering variable. 
Regression coefficients for valence and arousal are scaled per 10-point increase in the predictor.}
\end{minipage}
\end{threeparttable}
\end{table}

At the zero-order level, higher valence was positively associated with all social traits. The strongest correlations were observed for warmth, $\textit{r} = .57$, $p < .001$; friendliness, $\textit{r} = .48$, $p < .001$; likeability, $\textit{r} = .42$, $p < .001$; and trustworthiness, $\textit{r} = .31$, $p < .001$. Valence was also positively related to aliveness, $\textit{r} = .13$, $p < .001$; naturalness, $\textit{r} = .12$, $p < .001$; realness, $\textit{r} = .10$, $p = .001$; and humanlikeness, $\textit{r} = .08$, $p = .011$. Arousal was likewise positively associated with all social traits, with the strongest correlations observed for trustworthiness, $\textit{r} = .22$, $p < .001$; aliveness, $\textit{r} = .21$, $p < .001$; realness, $\textit{r} = .20$, $p < .001$; and humanlikeness, $\textit{r} = .19$, $p < .001$. All reported correlations remained significant after Holm correction.

When valence and arousal were entered simultaneously in the clustered GEE models, a more differentiated pattern emerged. Higher valence uniquely predicted greater warmth, $\textit{b} = 0.32$, $p < .001$; likeability, $\textit{b} = 0.21$, $p < .001$; trustworthiness, $\textit{b} = 0.12$, $p < .001$; friendliness, $\textit{b} = 0.28$, $p < .001$; and, to a lesser extent, aliveness, $\textit{b} = 0.05$, $p = .001$. In contrast, valence did not uniquely predict naturalness, $\textit{b} = 0.02$, $p = .124$; realness, $\textit{b} = 0.02$, $p = .166$; or humanlikeness, $\textit{b} = 0.02$, $p = .200$. Arousal showed the complementary pattern. Controlling for valence, higher arousal uniquely predicted greater aliveness, $\textit{b} = 0.14$, $p < .001$; naturalness, $\textit{b} = 0.10$, $p < .001$; realness, $\textit{b} = 0.10$, $p < .001$; and humanlikeness, $\textit{b} = 0.11$, $p < .001$. However, arousal did not uniquely predict warmth, $\textit{b} = -0.03$, $p = .282$; likeability, $\textit{b} = -0.02$, $p = .583$; trustworthiness, $\textit{b} = 0.02$, $p = .292$; or friendliness, $\textit{b} = -0.04$, $p = .135$. Overall, these continuous analyses suggest that valence was primarily related to socially positive impressions such as warmth, friendliness, likeability, and trustworthiness, whereas arousal is more strongly associated with impressions of aliveness, naturalness, realness, and humanlikeness.

\begin{figure}[h!]
    \centering
    \includegraphics[width=\linewidth]{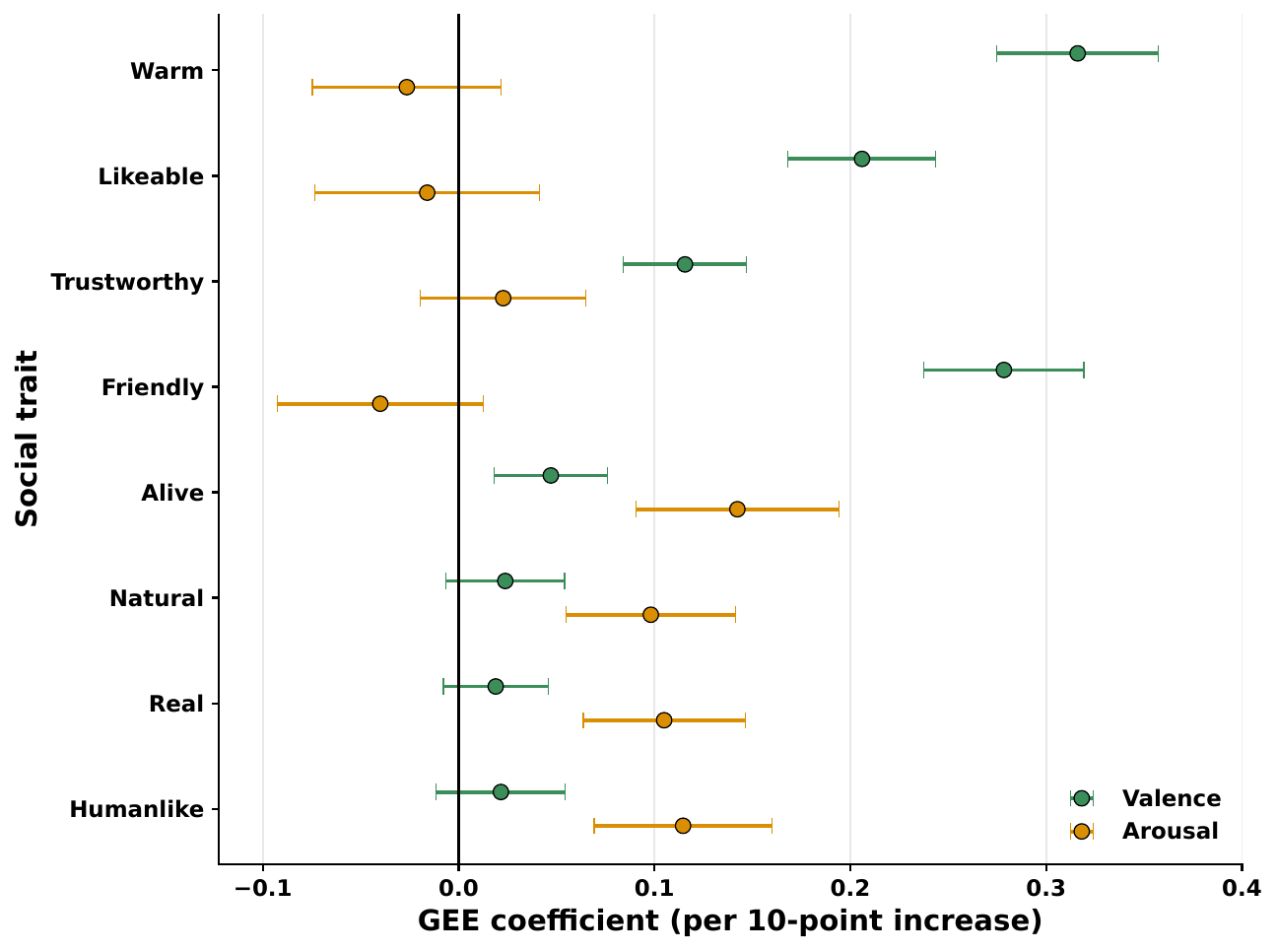}
\caption{Associations of social-trait ratings with continuous VA values, based on clustered GEE coefficients and 95\% CI.}
    \label{fig:ava}
\end{figure}

\subsection{Exploratory qualitative feedback}

Seventy-four participants provided optional open-ended comments at the end of the study. A descriptive review of these responses suggested three recurring patterns. First, participants frequently described emotion recognition as uneven: many reported that some expressions were easy to identify, whereas others were difficult, ambiguous, or required a ``best fit'' choice, particularly for subtler expressions. Second, comments about the robot itself were often positive, with many participants describing Reachy Mini as cute, endearing, or enjoyable to watch. Third, several participants explicitly referred to expressive or task-related constraints, including the absence of richer facial or eye cues, reliance on movement or voice to infer emotion, short clip duration, and the sense that some relevant labels were missing or overlapping in their opinion. Taken together, these qualitative observations align with the quantitative findings: the robot was generally perceived positively, emotion recognition was easier for some affective displays than for others, and was sometimes supported more by broader affective cues than by precise emotion labels.

\section{Discussion}

This study examined how people perceive emotional expressions displayed by Reachy Mini, a low-DoF robot with a constrained and clearly robot-specific expressive repertoire. Three main findings emerged. First, intended expressed emotions were not equally legible. Some expressions, especially anger, sadness, and interest, were recognized comparatively well, whereas others, such as love, pleasure, shame, and disgust, were much less reliably identified. This uneven pattern likely reflects the selective affordances of sparse, low-DoF expression, consistent with prior work showing that emotional readability in simplified robots depends strongly on embodiment and on which expressive channels are available \cite{StockHomburg2022,Ghafurian2022}. Second, broader affective meaning was often recovered more successfully than exact emotion labels. Participants were generally better at identifying valence- and arousal-level properties than the precise intended emotion, suggesting that Reachy Mini's expressions often communicated an approximate affective profile even when exact lexical agreement was weak. This suggests that simplified robot behavior may support coarse affective interpretation more robustly than fine-grained category recognition, in line with findings that people can extract emotional and social meaning even from abstract or non-humanoid robot movement \cite{Dubal2011,Embgen2012}. Third, emotional expressions shaped the social evaluation of the robot itself. Positive expressions were perceived as warmer and more sociable than negative ones, whereas animacy varied comparatively little across expressions. In bigger-picture terms, this indicates that robot affect functions not only as a signal to decode, but also as social information that shapes the robot's relational meaning, which is consistent with broader HRI accounts of emotion expression and social perception \cite{Henschel2021WhatYou,StockHomburg2022,Laban2024SharingFeel}.

These findings extend prior work showing that people can extract affective and social meaning from simplified or non-humanoid robot behavior \cite{Dubal2011,Embgen2012}, while also clarifying the limits of this capacity under strong expressive constraints. Rather than supporting uniform emotion recognition across the full set, Reachy Mini's low-DoF repertoire appears to preserve some affective distinctions more effectively than others. A central implication is that exact recognition and affective legibility should not be treated as equivalent outcomes. If evaluation relied only on exact-label accuracy, the robot's expressive capacity would appear quite limited. However, the quadrant-, valence-, and arousal-level results suggest a more nuanced account: even when participants did not converge on the intended label, they often still recovered the broader affective direction of the display. This analysis also revealed a dissociation between the roles of valence and arousal. Valence was primarily associated with socially evaluative traits such as warmth and likeability, whereas arousal was more strongly related to perceptions of animacy and realism. This indicates that different affective dimensions contribute to distinct aspects of social perception. For HRI, this distinction matters because many interactions may depend less on identifying a precise emotion term than on understanding whether the robot is signaling something positive or negative, calm or activated, or generally inviting versus aversive.

Several limitations should be acknowledged. The study was conducted online using short video clips rather than live interaction, and therefore reflects the perception of isolated displays rather than emotion expression embedded in reciprocal, situated encounters. Also, the use of GEW-inspired labels may have introduced lexical constraints for some subtle expressions, even though the broader affective analyses helped address this issue.  Future research will require comparison across additional robots and embodiments for making broader claims about low-DoF emotion expression.

%Overall, the findings suggest that Reachy Mini's emotional expressions are perceptible and socially consequential, but not uniformly so. The robot communicated some emotions relatively clearly and communicated broader affective structure more reliably than exact emotion labels across the full repertoire. These results support the value of evaluating robot emotion expression at multiple levels simultaneously and position Reachy Mini as a useful benchmark for studying affective communication in constrained robotic platforms.
\section{Conclusions}
This study showed that Reachy Mini's emotional expressions are perceptible and socially consequential, but not uniformly so. Some emotions were recognized relatively clearly, whereas others were less reliably identified, and broader affective structure was communicated more consistently than exact emotion labels across the full repertoire. The robot's expressions also shaped how warm and sociable it was perceived to be, even when participants did not identify the precise intended emotion. Taken together, these findings support the value of evaluating robot emotion expression at multiple levels simultaneously and position Reachy Mini as a useful benchmark for studying affective communication in constrained robotic platforms.

%\section*{Acknowledgment}

\bibliographystyle{IEEEtran}
\bibliography{references}

\end{document}